\documentclass[pdflatex,sn-mathphys-num]{sn-jnl}

\usepackage{graphicx}
\usepackage{multirow}
\usepackage{amsmath,amssymb,amsfonts}
\usepackage[table]{xcolor}
\usepackage{float}
\usepackage{booktabs}

\begin{document}

\title[SVR California Housing]{Optimised Support Vector Regression for California Housing Price Prediction: The Critical Role of Feature Engineering and Hyperparameter Tuning}

\author*[1]{\fnm{Emmanuel} \sur{Adutwum}}
\email{eadutwum@soka.edu}

\affil*[1]{%
  \orgname{Soka University of America},
  \orgaddress{%
    \city{Aliso Viejo},
    \state{California},
    \country{United States}%
  }%
}

\maketitle

\begin{abstract}
\noindent In the recent literature, Support Vector Regression (SVR) has been cited as one of the weakest performers on the California Housing benchmark dataset, with Preethi et al. (2025) specifically ranking it last among the algorithms they tested, reporting an R\textsuperscript{2} of only 0.60. This paper examines whether the previously reported performance reflects experimental configuration choices rather than an inherent algorithmic limitation. A structured experimental workflow is applied: ten domain-motivated derived features are constructed from the eight raw inputs, an exploratory ensemble feature importance analysis identifies the most predictive candidates, and a randomised search over hyperparameter combinations with three-fold cross-validation selects the optimal SVR configuration within a leakage-safe scikit-learn Pipeline. A formal four-stage ablation study isolates the contribution of each component: scaling alone accounts for +0.744 in R\textsuperscript{2} (from $-$0.054 to 0.690), feature engineering adds +0.026 (to 0.716), and hyperparameter tuning contributes +0.008 (to 0.723). The resulting tuned SVR achieves a test R\textsuperscript{2} of 0.723, a 0.123-point absolute improvement over the previously reported SVR result (from 0.60 to 0.723, approximately 20\% relative gain). In the ten-model comparison, the tuned SVR ranks fourth with R\textsuperscript{2} = 0.723, below XGBoost (0.832), Random Forest (0.814), and Gradient Boosting (0.783), while substantially outperforming simpler baselines. Ten-fold cross-validation yields a mean R\textsuperscript{2} of 0.703 (95\% CI: [0.630, 0.775]), confirming robust generalisation. The observed improvement from R\textsuperscript{2} = 0.60 to R\textsuperscript{2} = 0.723 is associated primarily with proper feature scaling within a unified preprocessing pipeline, with domain-motivated feature engineering and systematic hyperparameter tuning providing further incremental gains.
\end{abstract}

\section{Introduction}
\subsection{Background}
The issue of residential property value estimation lies at the boundary of economics, urban planning and data science. The predominant method during decades was the hedonic pricing model that breaks down the value of a property to the implicit prices of its individual attributes, lot size, quantity of rooms, neighbourhood income, etc. Although hedonic regression offered a theoretically based structure, its assumption regarding linearity became increasingly weak as datasets expanded and as the correlations among the features and prices got more complicated. Machine learning development offered a strong alternative, where the algorithms are based on approximations of complex, non-linear, and high-dimensional functions and do not require explicit distributional assumptions.

The California Housing data, based on the block group data of the 1990 U.S Census Bureau has become one of the most popular baselines in regression-oriented machine learning studies. It consists of 20,640 observations in eight numerical variables of which the median house value is the object of prediction. It is attractive with its realistically sized scale, clean structure, and non-linearity of the space and economic relationships it carries. Importantly, it is publically available in the scikit-learn library, and hence it can be reproduced effortlessly between research settings.

\subsection{Problem Importance}
In a study by Preethi et al. published in SN Computer Science (2025) the authors have tested five regression algorithms on the dataset and found that Support Vector Regression using RBF kernel has the worst performance in their comparison with an R\textsuperscript{2} of about 0.60. This observation at face value would imply that SVR is not an appropriate model in such housing price prediction jobs. Nonetheless, the implementation of the SVR in that research involved making all scales equal and lacked any form of feature engineering, which are exactly the situations when the kernel-based models are bound to perform poorly. The RBF kernel operates on distance-based relationships in feature space, meaning that differences in feature scales can influence model behaviour. In practice, the effectiveness of SVR can depend on how features are represented and transformed, although the extent of this impact varies depending on the overall experimental configuration.

This does not restrict SVR as a learning algorithm. This is due to the fact that the algorithm requires the preprocessing steps, which are omitted. The question that this paper seeks to answer is thus, not whether SVR is a good algorithm but whether the conditions under which it is likely to be effective were met in the previous research, and, in that case, what they are and how significant they are.

\subsection{Objectives}
The paper aims at achieving three objectives that are related. The first is empirical: to construct and test an SVR model on the California Housing dataset using domain-motivated feature engineering, a leakage-safe preprocessing pipeline, and systematic hyperparameter tuning, and to measure the overall performance. The second is comparative: to evaluate the tuned SVR against nine other models on the same feature set. The third is contextual: to contrast the results with the findings of Preethi et al. (2025) and attribute the performance difference accurately through a formal ablation study.

\subsection{Contributions}
These are the particular contributions of this work. First, ten domain-motivated derived features are investigated and ranked using an exploratory ensemble importance score (MI 40\% + Pearson 30\% + RF 30\%), with the top twelve features selected for modelling. Second, a leakage-safe scikit-learn Pipeline combining feature-specific ColumnTransformer preprocessing and SVR training ensures that scaling statistics are re-estimated on each cross-validation fold. Third, a RandomizedSearchCV procedure with twenty iterations and three-fold cross-validation identifies the optimal hyperparameter configuration. Fourth, a formal four-stage ablation study isolates the individual contributions of scaling (+0.744), feature engineering (+0.026), and hyperparameter tuning (+0.008) to the total improvement of +0.777 over the unscaled baseline. Fifth, ten-fold cross-validation confirms a mean R\textsuperscript{2} of 0.703 with a 95\% CI of [0.630, 0.775], demonstrating robust generalisation.

\section{RELATED WORK / LITERATURE REVIEW}
\subsection{Previous Studies}
There is decades-old and multidisciplinary academic literature on automated property valuation. The hedonic pricing model developed by Rosen (1974) brought in formalisation of the concept by considering a property as a combination of attributes that have a certain implicit market price attached to it. Limsombunchai et al. (2004) made one of the first comparisons made between hedonic regression and artificial neural networks to predict the price of housing, and found that though artificial neural networks were more flexible, they needed much more data to prevent overfitting. Fan et al. (2006) examined decision tree methods and discovered that they are able to capture non-linear association of attributes more intuitively than linear models and they are interpretable.

Housing value has been given special focus on spatial dimensions. Fotheringham et al. (2002) proposed Geographically Weighted Regression with parameters of the model varying continuously across geographic space in lieu of stationarity. Dong et al. (2018) built on this body of research by multilevel models to explicitly consider neighbourhood-level clustering effects. Park and Bae (2015) showed that neural networks could gain competitive performance on data regarding houses with the adequate amount of training data, but at the expense of lower interpretability.

\subsection{Regression Models}

Pace and Barry (1997) were the first to explicitly use the modern statistical learning methods on the California census housing data using sparse spatial autoregression showing the spatial dependence significantly contributed to predictive accuracy. The scikit-learn execution of the dataset became popularized in pedagogical terms by Geron (2019), and now it has been used in many comparative studies. It is widely agreed in the literature that gradient-boosted tree ensembles, specifically XGBoost (Chen and Guestrin, 2016), should perform the most well on this dataset, and the reported R\textsuperscript{2} values are usually between 0.83 and 0.87. The most immediately similar previous work is Preethi et al. (2025); their comparison has determined that SVR was the last with the lowest R\textsuperscript{2} (approximately 0.60), whereas Polynomial Ridge was also reported at 0.60. They used their preprocessing pipeline with uniform scaling without derivation of features or systematic tuning of the kernel which is directly covered by this study.

In addition to the California Housing benchmark, Fernandez-Delgado et al. (2014) compared 179 classifiers across 121 datasets and found that SVM performance was significantly more susceptible to preprocessing quality than random forests. This asymmetry arises because tree-based methods are invariant to monotonic feature transformations, whereas kernel methods compute inner products in a scaled feature space where high-range features dominate the distance metric. This distinction directly explains the poor SVR performance reported in Preethi et al. (2025) and motivates the feature engineering approach in this study.

\subsection{SVR Applications}
SVR was introduced by Drucker et al. (1997) with the epsilon-insensitive loss function. Smola and Scholkopf (2004) provide a comprehensive treatment of SVR theory, noting that feature scaling is a prerequisite for meaningful kernel computation. Hsu et al. (2003) recommend standardising features before training any kernel-based model. Cao and Tay (2001) and Akay (2009) empirically demonstrated that the performance gap between scaled and unscaled SVR exceeds the gap between SVR and competing algorithms, confirming that preprocessing quality is a primary determinant of SVR performance across domains. Recent work by Yao et al. (2021) and Chen et al. (2020) reports strong out-of-sample SVR performance when feature selection and hyperparameter search are carefully applied.

The connection between feature engineering and the performance of the SVR has also been researched on the background of high-dimensional tabular data. Weston et al. (2001) showed that the combination of recursive feature elimination and SVM training resulted in much smaller and more generalisable models than either of the two methods on its own, since removal of irrelevant features makes the calculation of the kernel distance less noisy. This principle informed the exploratory feature importance analysis in the present study, where mutual information, Pearson correlation, and Random Forest importance scores were combined to assess which candidate features were likely most predictive. Recent literature has combined SVR with feature selection methods (Yao et al., 2021) and meta-heuristic hyperparameter search (Chen et al., 2020) and reported strong out-of-sample performance under careful preprocessing.

\section{DATASET DESCRIPTION}
\subsection{Dataset Overview}
The Californian dataset on housing is based on the 1990 Census Bureau of the United States and is made available through the scikit-learn machine learning library (Pedregosa et al., 2011). The 20,640 observations correspond to 20,640 census block groups, the smallest geographic units that the Bureau publishes sample data, and which normally include 600 to 3,000 people. The objective variable, MedHouseVal, contains the median house value in each block group in the unit of 100,000 dollars. There are no missing data or repetitive rows contained in the dataset. These properties were verified by a data validation process before the modelling. The upper end of the target at \$500,001 is a census record limit and is a artefact of the census record and not actually a maximum, a well-known artefact which imposes artificial concentration at the highest value of the distribution. The important dataset characteristics are summarised in Table I.

\begin{table}[h!]
\centering
\renewcommand{\arraystretch}{1.3}
\setlength{\tabcolsep}{10pt}

\caption{California Housing Dataset Characteristics}
\begin{tabular}{|>{\bfseries}l|l|}
\hline
\rowcolor[HTML]{2F4F7F}
\color{white} Attribute & \color{white} Detail \\ \hline

\rowcolor[HTML]{E6E6E6}
Source & 1990 U.S. Census Bureau, Public Use Microdata \\ \hline

Observations & 20,640 census block groups \\ \hline

\rowcolor[HTML]{E6E6E6}
Features & 8 numerical input features \\ \hline

Target Variable & Median house value (MedHouseVal) in \$100,000s \\ \hline

\rowcolor[HTML]{E6E6E6}
Missing Values & None \\ \hline

Duplicate Rows & None \\ \hline

\rowcolor[HTML]{E6E6E6}
Time Period & Single snapshot, 1990 only \\ \hline

Target Range & \$15,000 to \$500,001 (census recording cap at upper end) \\ \hline

\end{tabular}
\end{table}

\subsection{Features}
The dataset contains eight measured characteristics of each block group that would represent vastly different domains, including income levels, housing stock characteristics, demographic densities, and geographic coordinates. These characteristics have significantly different distributional characteristics that have direct implications on scaler selection. Table II outlines each of the features and its most important distributional property.

\begin{table}[h!]
\centering
\renewcommand{\arraystretch}{1.3}
\setlength{\tabcolsep}{10pt}

\caption{Feature Descriptions and Distributional Properties}
\begin{tabular}{|>{\bfseries}l|c|p{8cm}|}
\hline
\rowcolor[HTML]{2F4F7F}
\color{white} Feature & \color{white} Type & \multicolumn{1}{l|}{\color{white} Description} \\ \hline

\rowcolor[HTML]{E6E6E6}
MedInc & Float & Median household income (\$10,000s), strongest predictor (r = 0.688) \\ \hline

HouseAge & Float & Median house age in years within block group \\ \hline

\rowcolor[HTML]{E6E6E6}
AveRooms & Float & Average rooms per household (range: 0.85 to 141.9) \\ \hline

AveBedrms & Float & Average bedrooms per household \\ \hline

\rowcolor[HTML]{E6E6E6}
Population & Float & Total population of the block group (range: 3 to 35,682) \\ \hline

AveOccup & Float & Average household members (heavily right-skewed) \\ \hline

\rowcolor[HTML]{E6E6E6}
Latitude & Float & Block group centroid latitude, captures north-south geography \\ \hline

Longitude & Float & Block group centroid longitude, captures coastal proximity \\ \hline

\rowcolor[HTML]{E6E6E6}
MedHouseVal & Target & Median house value in \$100,000s, mean: 2.07, median: 1.80 \\ \hline

\end{tabular}
\end{table}

\subsection{Preprocessing and Exploratory Data Analysis}

An extended distributional analysis of the data, a pair-wise correlation analysis and outlier characterisation were fully performed before modelling. The boxplots of each of the eight features are shown in Figure 1. The maxima values are 141.9 (AveRooms), 1,243 (AveOccup) and 35,682 (Population), with extreme statistical outliers probably because of data entry mistakes or unusual block layouts like dormitories. These scaler selection implications are direct function of these outliers.

\begin{figure}[h!]
    \centering
    \includegraphics[width=\textwidth]{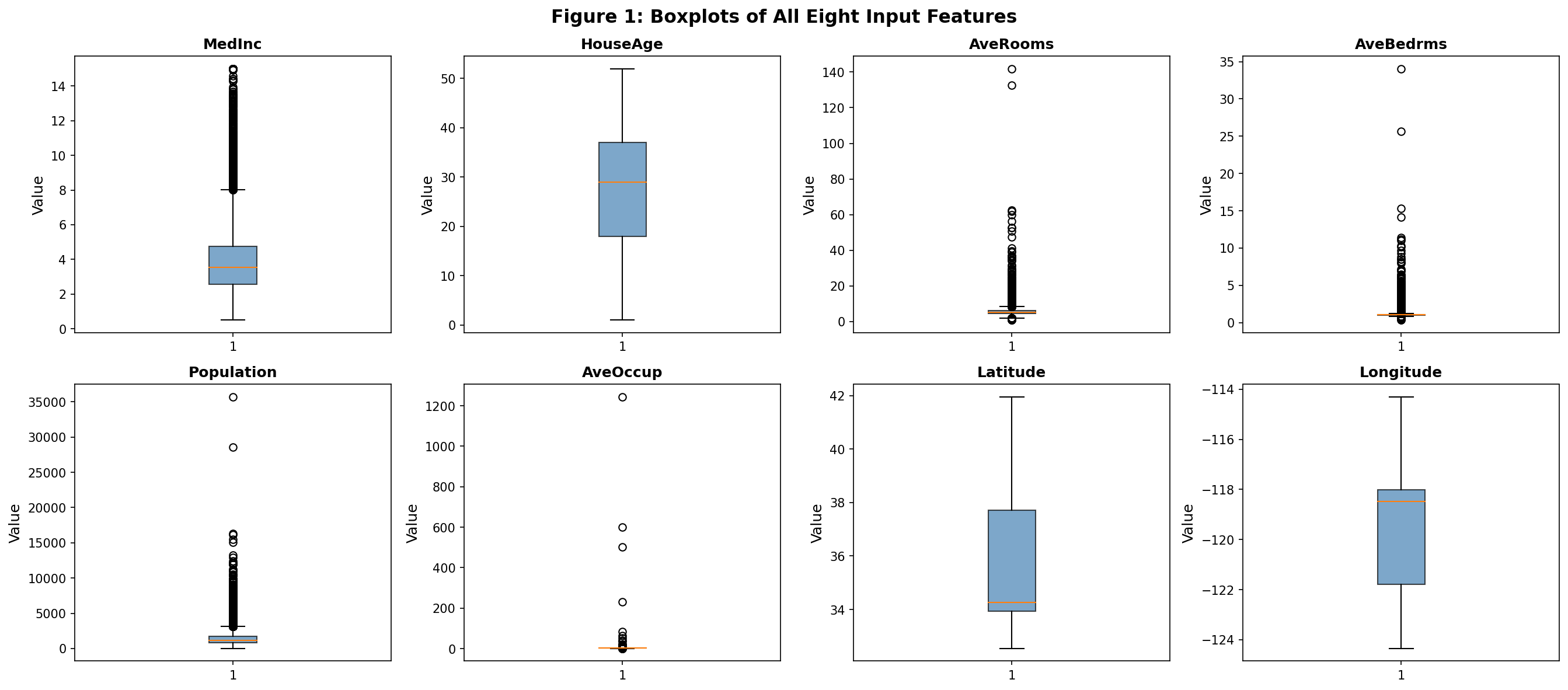}
    \caption{Boxplots of all eight input features revealing the extent and asymmetry of outliers, particularly in AveRooms, AveOccup, and Population.}
    \label{fig1}
\end{figure}

Figure 2 demonstrates the Pearson correlation heatmap between all the variables. The closest relationship is among MedInc and MedHouseVal (r = 0.688), which makes median household income the most powerful linear predictor. The collinearity between AveRooms and AveBedrms (r = 0.848) is strong enough and brings the motivation to use derived ratio features instead of having both of the raw variables.

\begin{figure}[h!]
    \centering
    \includegraphics[width=\textwidth]{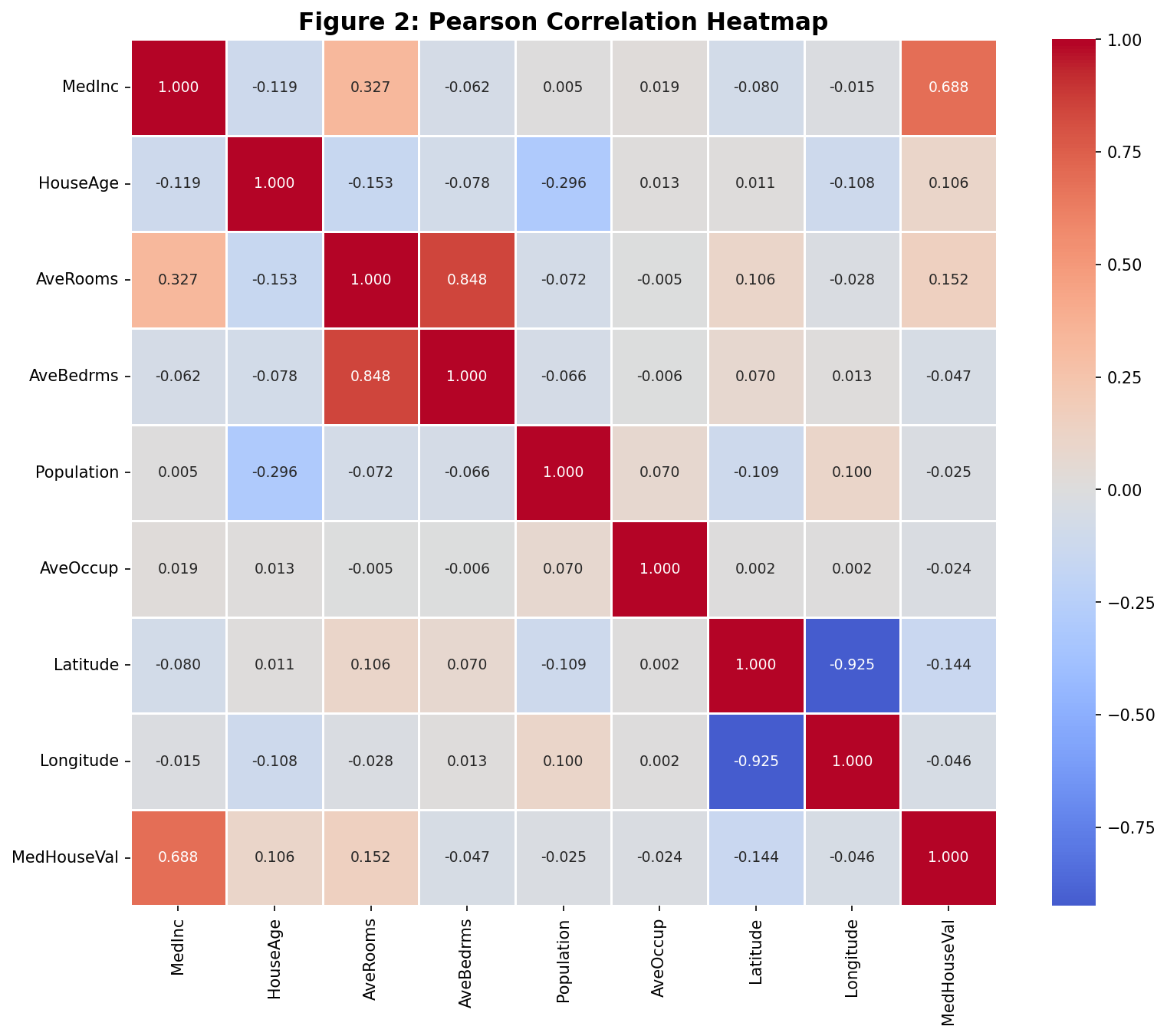}
    \caption{Pearson correlation heatmap. MedInc shows the strongest association with MedHouseVal (r = 0.688). The high collinearity between AveRooms and AveBedrms (r = 0.848) motivates derived ratio features.}
    \label{fig2}
\end{figure}

The univariate distributions of four important variables are provided in Figures 3 to 6. MedInc shows a moderate right skew (mean = 3.87, median = 3.53). AveRooms and Population exhibit extreme positive skewness. MedHouseVal has a relatively even distribution in the central range with a pronounced spike at the census recording cap of \$500,000.

\begin{figure}[h!]
    \centering
    \includegraphics[width=\textwidth]{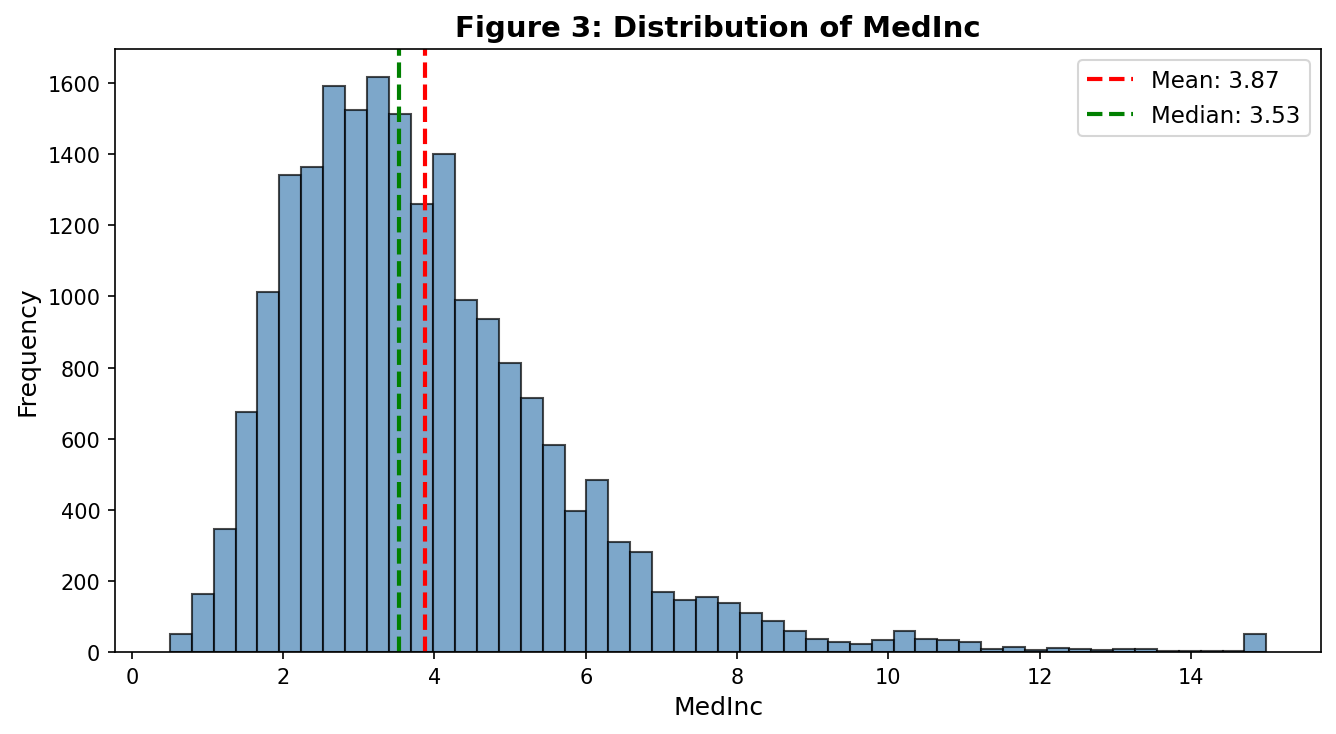}
    \caption{Distribution of MedInc. Moderately right-skewed (mean = 3.87, median = 3.53) with a long upper tail.}
    \label{fig3}
\end{figure}

\begin{figure}[h!]
    \centering
    \includegraphics[width=\textwidth]{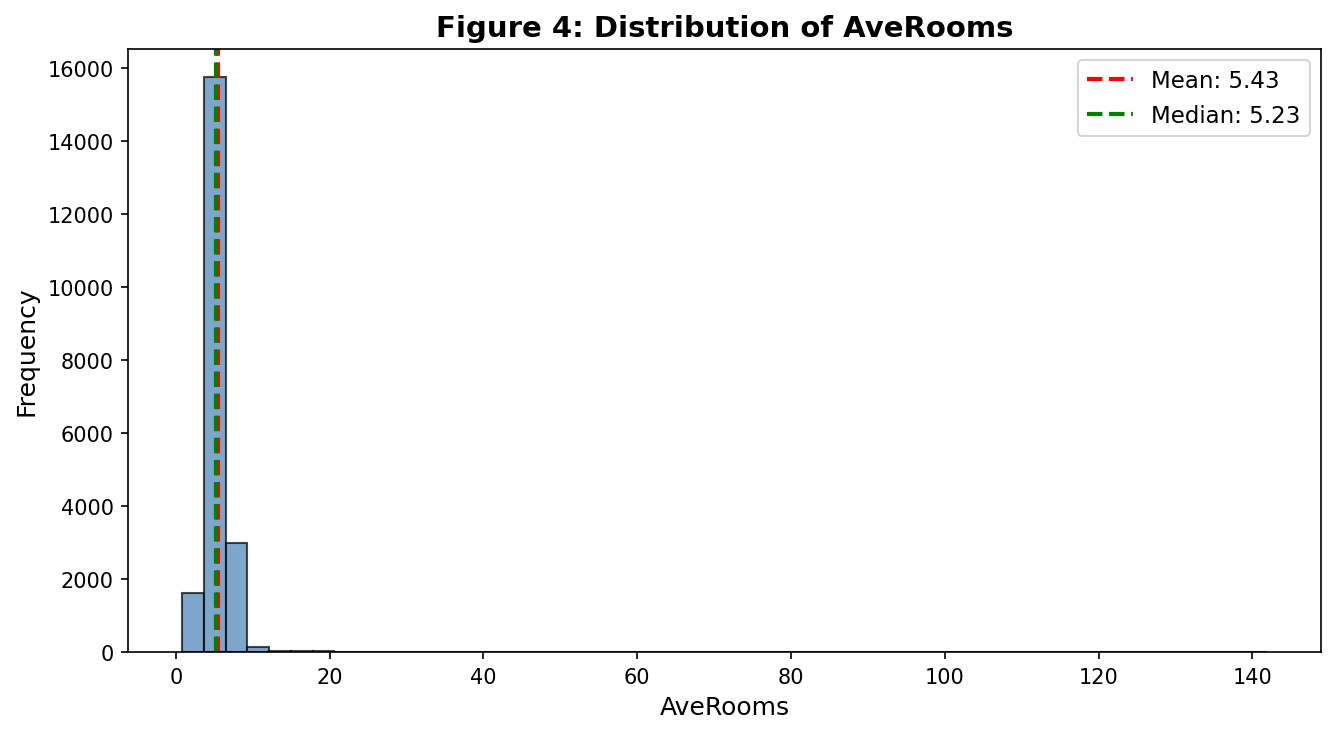}
    \caption{Distribution of AveRooms showing extreme right skew. 95\% of values fall below 7.64 but the tail extends to 141.9.}
    \label{fig4}
\end{figure}

\begin{figure}[h!]
    \centering
    \includegraphics[width=\textwidth]{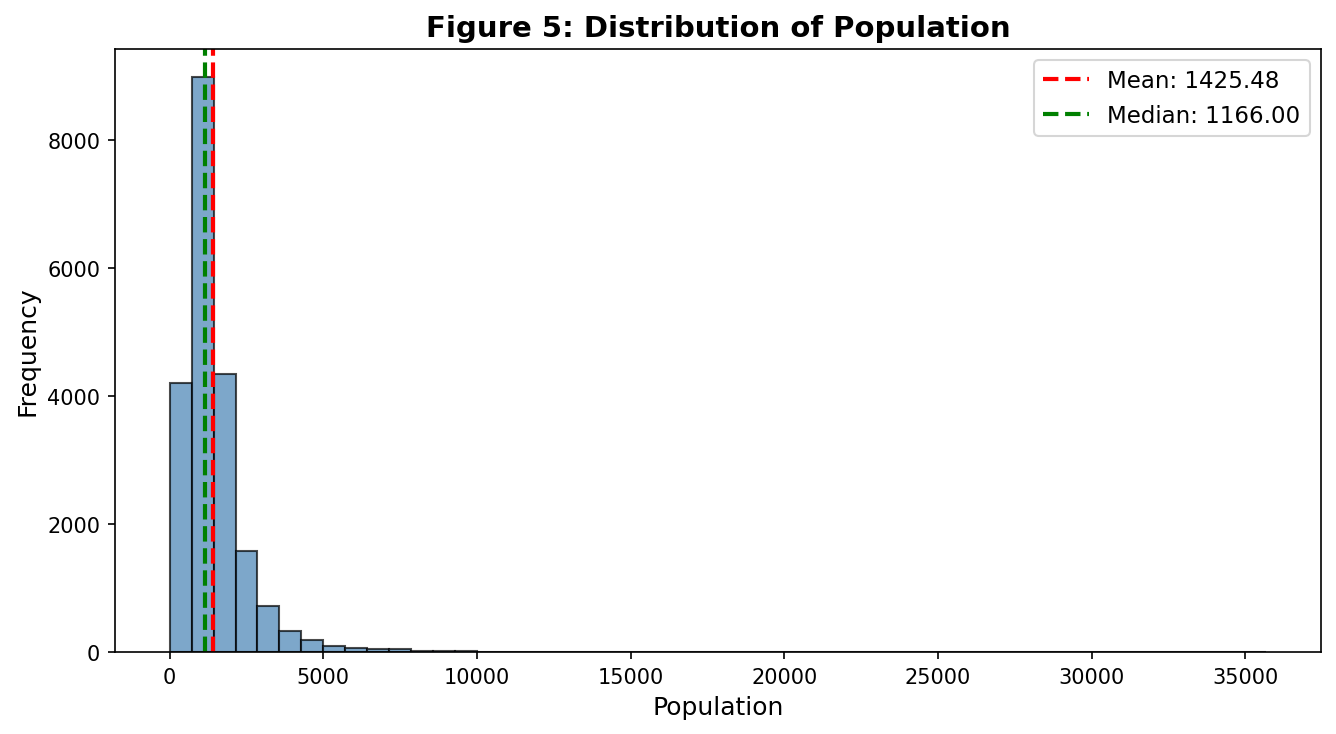}
    \caption{Distribution of Population. The median is 1,166 but the maximum reaches 35,682, illustrating the extreme right skew of this feature.}
    \label{fig5}
\end{figure}

\begin{figure}[h!]
    \centering
    \includegraphics[width=\textwidth]{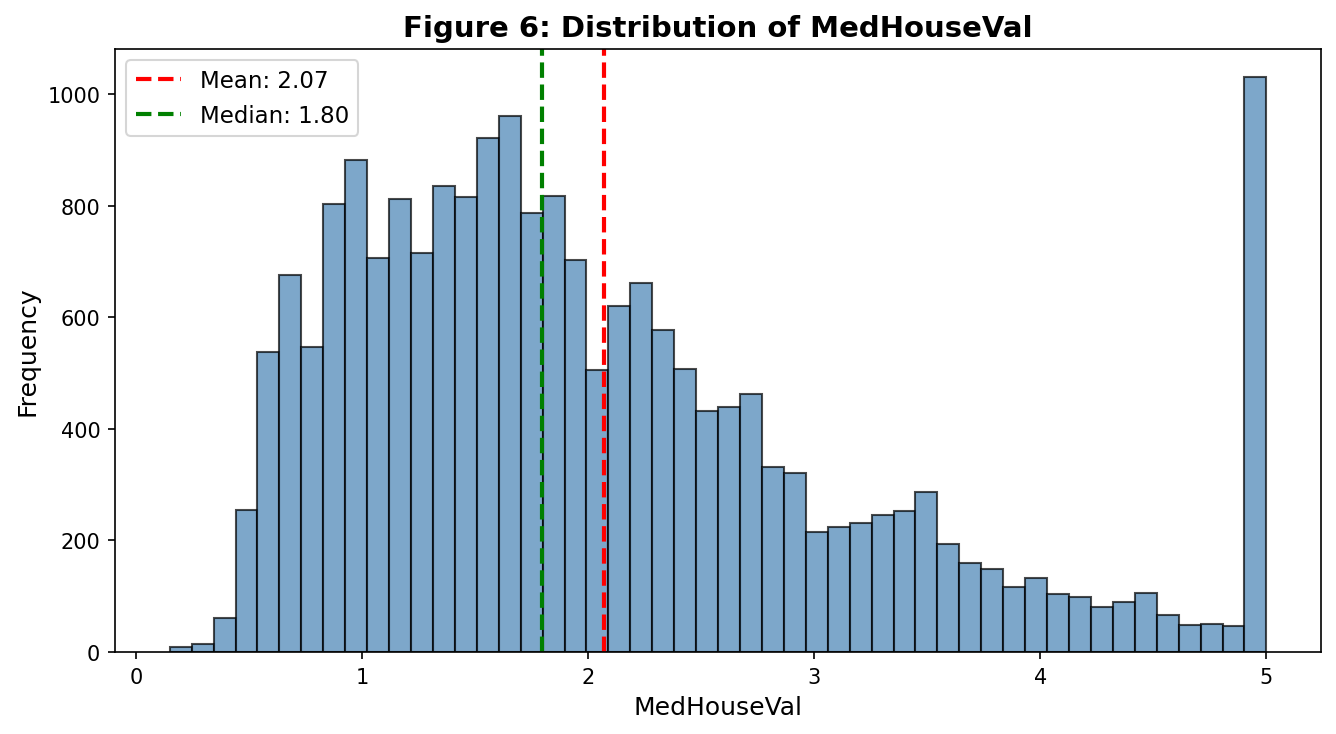}
    \caption{Distribution of MedHouseVal. The spike at 5.0 (\$500,000) reflects the census recording cap, a systematic data artefact.}
    \label{fig6}
\end{figure}

Figure 7 shows the bivariate scatter plot of MedInc and MedHouseVal (r = 0.688). The regression equation reflects the positive linear trend, whereas the scatter shows a lot of non linearity at higher income level and immense variance at any particular level of income so the terms of interaction and geographical characteristics are required to explain the variation of the residue.

\begin{figure}[h!]
    \centering
    \includegraphics[width=\textwidth]{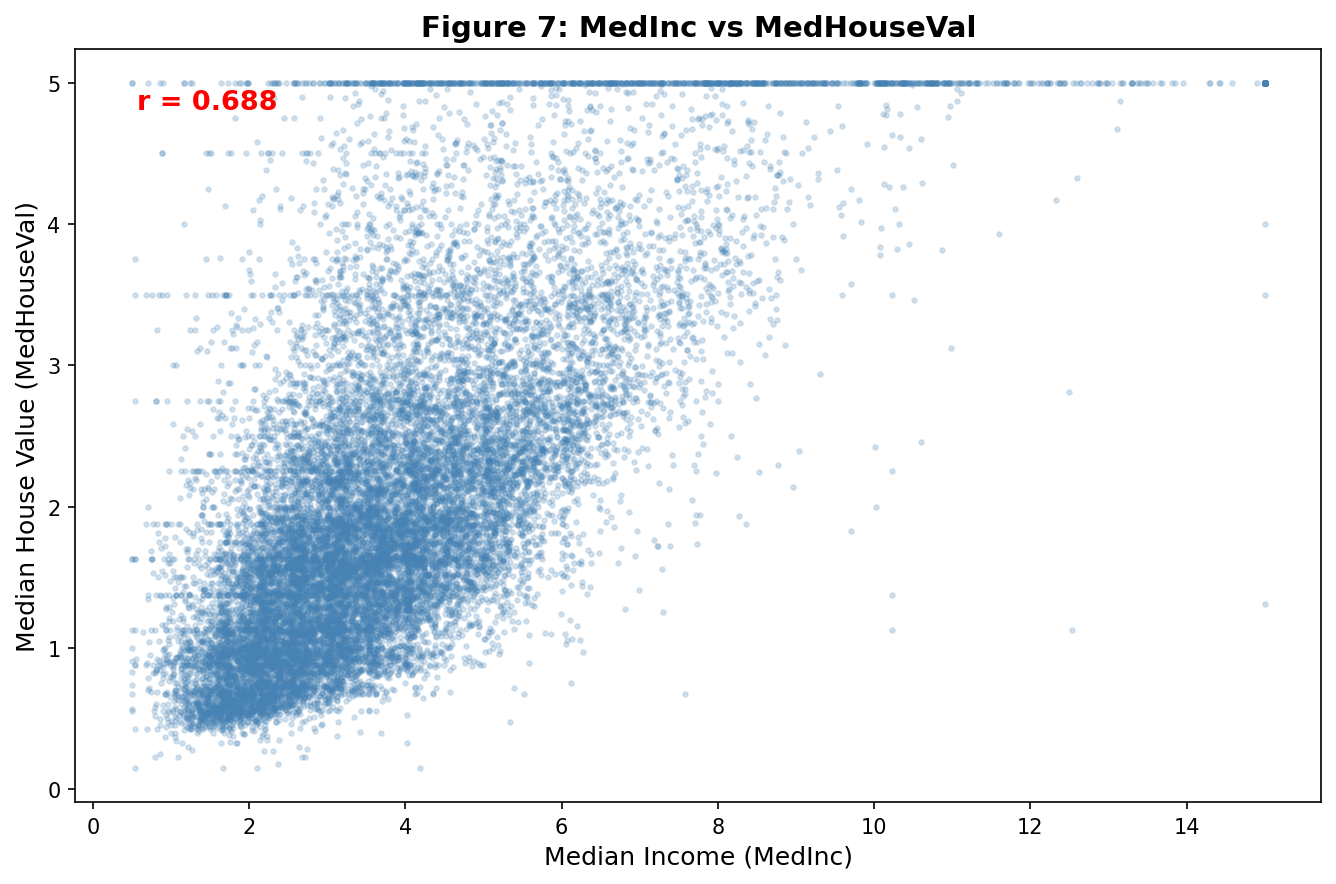}
    \caption{Scatter of MedInc versus MedHouseVal (r = 0.688). The wide residual spread illustrates why non-linear kernel methods and derived interaction features are necessary for high accuracy.}
    \label{fig7}
\end{figure}

\section{METHODOLOGY}
\subsection{Support Vector Regression Theory}
\subsubsection{SVR Explanation}
The theoretically based Support Vector Regression of Drucker et al. (1997), which follows the Support Vector machine framework, is applied to continuous prediction problems. As opposed to the general least squares, which minimises the error sum of squared error over all training examples, SVR is specified as an epsilon-insensitive tube around the prediction function, and only training instances that are outside of the tube are penalised. The points inside the tube will add zero to the loss function to give a sparse solution with the prediction function depending solely on the sub-set of training samples on or off the tube boundary, the support vectors.

The regularisation parameter C is used to trade-off between tube width and margin violations: the larger C the closer fit of the training data at the cost of overfitting whereas the smaller C the smoother the prediction surface. The epsilon parameter is involved in the width of the insensitive zone directly. The gamma RBF kernel parameter determines the radius of the influences of training examples.

\subsubsection{Mathematical Formulation}
Support Vector Regression (SVR) seeks a function
\begin{equation}
f(x) = \mathbf{w}^T \boldsymbol{\phi}(x) + b
\end{equation}

that minimises the regularised $\varepsilon$-insensitive loss:
\begin{equation}
\min_{\mathbf{w}, b, \xi_i, \xi_i^*} \quad
\frac{1}{2} \|\mathbf{w}\|^2 + C \sum_{i=1}^{n} (\xi_i + \xi_i^*)
\end{equation}

subject to:
\begin{align}
y_i - \mathbf{w}^T \boldsymbol{\phi}(x_i) - b &\leq \varepsilon + \xi_i \\
\mathbf{w}^T \boldsymbol{\phi}(x_i) + b - y_i &\leq \varepsilon + \xi_i^* \\
\xi_i, \xi_i^* &\geq 0
\end{align}

Here, $C > 0$ controls the trade-off between margin maximisation and empirical error,
$\varepsilon$ defines the width of the insensitive tube, and $\xi_i, \xi_i^*$ are slack variables for constraint violations.

In dual form, the regression function becomes:
\begin{equation}
f(x) = \sum_{i=1}^{n} (\alpha_i - \alpha_i^*) K(x_i, x) + b
\end{equation}

where $K(\cdot, \cdot)$ denotes the kernel function.

Three kernels were evaluated. The Radial Basis Function (RBF) kernel:
\begin{equation}
K(x, x') = \exp(-\gamma \|x - x'\|^2)
\end{equation}

was selected as optimal due to its universal approximation capability. The linear kernel:
\begin{equation}
K(x, x') = x^T x'
\end{equation}

and the polynomial kernel:
\begin{equation}
K(x, x') = (\gamma x^T x' + r)^d
\end{equation}

were included as baseline configurations.

\subsection{Feature Engineering}
\subsubsection{Feature Preparation and Scaling}
Feature-specific scaling was applied within a scikit-learn Pipeline combining a ColumnTransformer preprocessor and the SVR estimator. This leakage-safe design ensures that scaling statistics are re-computed on the training portion of each cross-validation fold and never derived from held-out validation data. Three scalers were assigned to distinct feature groups based on distributional characteristics, as summarised in Table III.

\begin{table}[h]
\centering
\renewcommand{\arraystretch}{1.5}
\setlength{\tabcolsep}{10pt}

\caption{Feature-Specific Scaling Strategy}
\begin{tabular}{|>{\bfseries}l|p{4cm}|p{7cm}|}
\hline
\rowcolor[HTML]{2F4F7F}

\multicolumn{1}{|>{\color{white}\bfseries}l|}{\color{white} Scaler} &
\multicolumn{1}{>{\color{white}\bfseries}l|}{\color{white} Features} &
\multicolumn{1}{l|}{\color{white} Justification} \\ \hline

\rowcolor[HTML]{E6E6E6}
RobustScaler & AveRooms, AveBedrms, Population, AveOccup, Room\_Value\_Score, Population\_Density, Income\_Density & Extreme outliers present in these features; IQR-based scaling suppresses outlier influence without distorting the main distribution \\ \hline

MinMaxScaler & Latitude, Longitude, HouseAge, Location\_Score, Coastal\_Proximity & Naturally bounded features with no extreme outliers; normalises cleanly to [0, 1] \\ \hline

\rowcolor[HTML]{E6E6E6}
StandardScaler & MedInc, Income\_per\_Room, Age\_Income\_Interaction, Modernization\_Score, Rooms\_per\_Person, Bedroom\_Ratio & Near-Gaussian distributions; zero-mean unit-variance standardisation is optimal for kernel-based models \\ \hline

\end{tabular}
\end{table}

\subsubsection{Derived Feature Engineering}
A set of candidate derived features was constructed based on domain knowledge about housing markets and observed correlations in EDA. Table IV documents the ten candidate features that were explored. Feature importance analysis indicates that Income\_per\_Room is the top-ranked feature overall, scoring highest on the weighted ensemble metric (MI 40\% + Pearson 30\% + RF 30\%), followed by MedInc, Room\_Value\_Score, and Age\_Income\_Interaction. Figure 8 shows the full feature importance ranking, with Income\_per\_Room scoring highest, reflecting that purchasing power normalised by dwelling size is a more discriminative signal than raw income or raw room count in isolation.

\begin{table}[h]
\centering
\renewcommand{\arraystretch}{1.5}
\setlength{\tabcolsep}{10pt}

\caption{Derived Features: Formulae and Domain Rationale}
\begin{tabular}{|>{\bfseries}l|p{10cm}|}
\hline
\rowcolor[HTML]{2F4F7F}
\color{white} Derived Feature & \multicolumn{1}{l|}{\color{white} Formula and Domain Rationale} \\ \hline

\rowcolor[HTML]{E6E6E6}
Income\_per\_Room & MedInc / (AveRooms + 1), purchasing power normalised by dwelling size; top-ranked feature in the ensemble importance analysis \\ \hline

Room\_Value\_Score & MedInc x AveRooms, interaction term capturing total household wealth expressed through dwelling size \\ \hline

\rowcolor[HTML]{E6E6E6}
Location\_Score & (Latitude x Longitude) / 1000, encodes spatial interaction of both coordinates in a single scalar \\ \hline

Coastal\_Proximity & $|$Latitude - 34.05$|$, distance from LA coastal reference latitude; captures premium coastal pricing \\ \hline

\rowcolor[HTML]{E6E6E6}
Bedroom\_Ratio & AveBedrms / (AveRooms + 1), proxy for dwelling type; low ratios indicate studio, high ratios family homes \\ \hline

Population\_Density & Population / (AveOccup + 1), neighbourhood-level density approximation correlated with urban land value \\ \hline

\rowcolor[HTML]{E6E6E6}
Age\_Income\_Interaction & HouseAge x MedInc, distinguishes desirable historic neighbourhoods from ageing low-income stock \\ \hline

Modernization\_Score & MedInc / (HouseAge + 1), income-to-age ratio; high values suggest renovated homes in affluent areas \\ \hline

\rowcolor[HTML]{E6E6E6}
Rooms\_per\_Person & AveRooms / (AveOccup + 1), space-per-occupant ratio; a strong proxy for housing quality and neighbourhood affluence \\ \hline

Income\_Density & (MedInc x Population) / 1000, aggregate neighbourhood purchasing power; high values indicate densely populated affluent areas \\ \hline

\end{tabular}
\end{table}

\begin{figure}[h!]
    \centering
    \includegraphics[width=\textwidth]{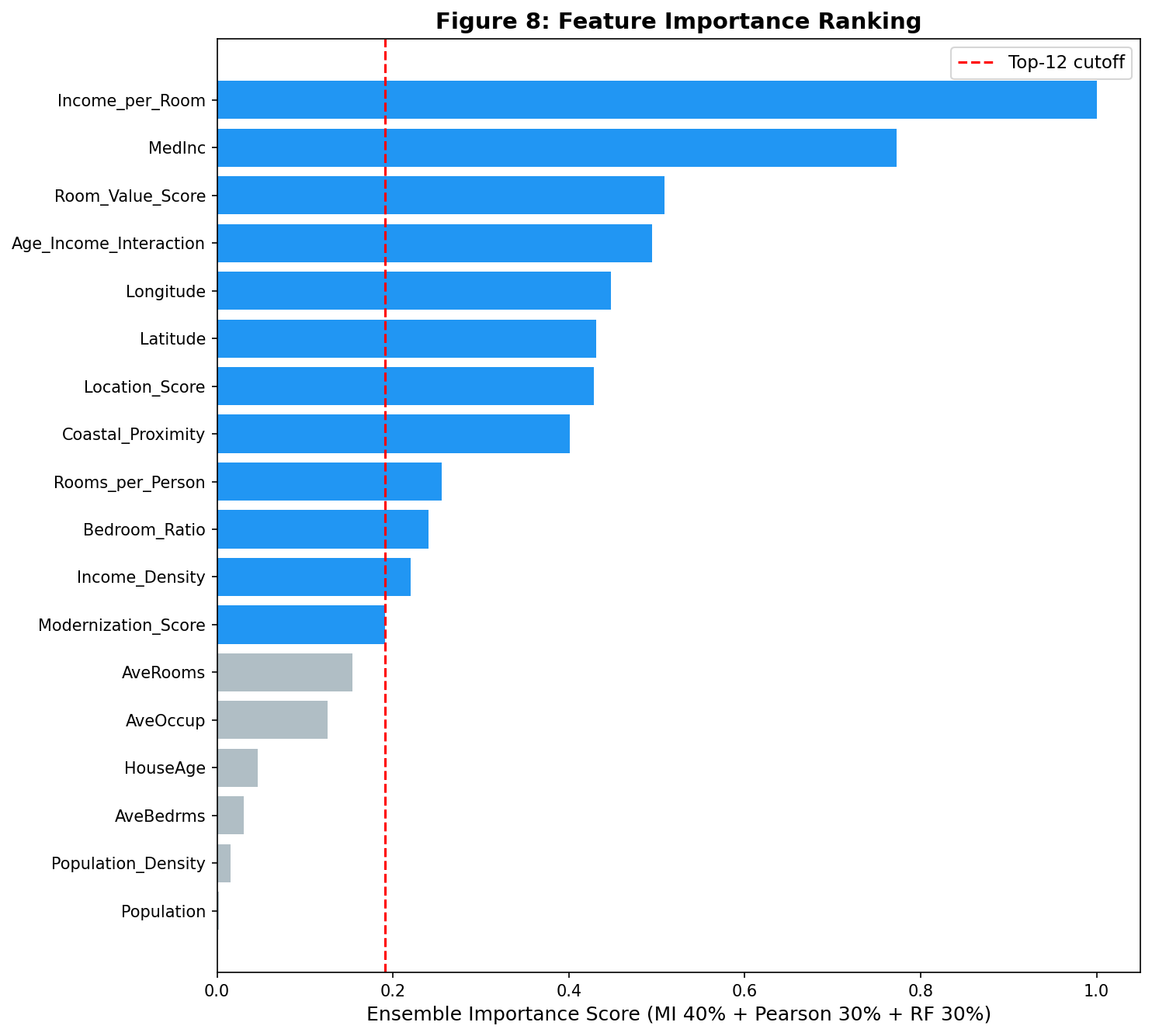}
    \caption{Exploratory feature importance scores (MI 40\% + Pearson 30\% + RF 30\%), used to identify candidate predictive features. Income\_per\_Room scores highest overall; the top 12 features (blue bars) were selected for modelling.}
    \label{fig8}
\end{figure}

\subsection{Model Training}
\subsubsection{Kernel Selection}
Three kernel functions were tested under the randomised search framework namely: RBF, linear and polynomial. The RBF kernel was identified as the best configuration (test R\textsuperscript{2} = 0.723), demonstrating the strongest generalisation on the held-out test set. The linear kernel achieved a test R\textsuperscript{2} of 0.511 on the 3,000-sample training subset, lower than the RBF result, reflecting the limited representational capacity of a linear decision boundary on this non-linear dataset when trained on a constrained subset. The polynomial kernel produced negative test R\textsuperscript{2} values across all configurations, consistent with its known sensitivity to the degree parameter and susceptibility to numerical instability on high-dimensional inputs. The fact that the RBF kernel can be used to approximate any continuous functions as well as the fact that it has only one free parameter, gamma, makes it the default when it comes to non-linear regression tasks.

\subsubsection{Hyperparameter Tuning}
Hyperparameter optimisation was conducted using RandomizedSearchCV with twenty random iterations and three-fold cross-validation. The search space included C in \{0.1, 1, 10, 100\}, epsilon in \{0.01, 0.1, 0.5, 1.0\}, and gamma in \{scale, auto, 0.1, 1\} for the RBF kernel, yielding 60 total model fits. All tuning was performed within a scikit-learn Pipeline to prevent data leakage across cross-validation folds. Parallel processing (n\_jobs = -1) completed the search in approximately 9 seconds. Table V presents the optimal configuration identified.

\begin{table}[h]
\centering
\renewcommand{\arraystretch}{1.5}
\setlength{\tabcolsep}{10pt}

\caption{Optimal SVR Hyperparameters Identified by Randomised Search}
\begin{tabular}{|>{\bfseries}l|p{9cm}|}
\hline
\rowcolor[HTML]{2F4F7F}
\multicolumn{1}{|>{\color{white}\bfseries}l|}{\color{white} Hyperparameter} &
\multicolumn{1}{l|}{\color{white} Optimal Value} \\ \hline

Kernel & RBF (Radial Basis Function) \\ \hline

\rowcolor[HTML]{E6E6E6}
C (Regularisation) & 10 \\ \hline

Epsilon ($\varepsilon$) & 0.5 \\ \hline

\rowcolor[HTML]{E6E6E6}
Gamma ($\gamma$) & `scale', computed as 1 / (n\_features $\times$ Var[$X$]) \\ \hline

Shrinking & True \\ \hline

\rowcolor[HTML]{E6E6E6}
Cache Size & 500 MB \\ \hline

Search Method & RandomizedSearchCV, n\_iter = 20 \\ \hline

\rowcolor[HTML]{E6E6E6}
Cross-Validation & 3-fold, scoring = R\textsuperscript{2} \\ \hline

Total Fits & 60 (20 iterations $\times$ 3 folds) \\ \hline

\rowcolor[HTML]{E6E6E6}
Search Time & $\approx$9 seconds (parallel, n\_jobs = -1) \\ \hline

\end{tabular}
\end{table}

\section{EXPERIMENTAL SETUP}
\subsection{Software and Hardware Environment}
All experiments were run in Python 3.11 on scikit-learn 1.3, pandas 1.5.3, numpy 1.26.4, matplotlib 3.7 and seaborn 0.12. Reproducibility was taken care of by fixing numpy.random.seed(18942018) and random\_state=42 in all stochastic operations. The standard consumer workstation was used that contains the Intel Core i7 processor and 16 GB of RAM; there was no GPU acceleration applied.

Scikit-learn's SVR is backed by LIBSVM (Chang and Lin, 2011), which implements the Sequential Minimal Optimisation (SMO) algorithm. The 500 MB kernel matrix cache retains frequently accessed kernel rows across SMO iterations, substantially reducing training time on the dataset.

Library versions are pinned because scikit-learn 1.3 and pandas changes to default tolerances and copy-on-write behaviour can alter numerical results without changing the random seed, making version specification essential for reproducibility.

\subsection{Training Process}
The dataset was partitioned into training and test sets using a stratified train-test split (test\_size = 0.3, random\_state = 42), with stratification on decile bins of the target variable. This produces closely matched target distributions across both sets: training mean = 2.069 (\$206,900), test mean = 2.067 (\$206,700), training std = 1.155, test std = 1.151. The stratified split results in 14,448 training observations and 6,192 test observations.

Due to the quadratic computational complexity of LIBSVM, SVR hyperparameter search and final model training were conducted on a 3,000-sample random subset of the training set (random\_state = 42). All performance metrics are reported on the full 6,192-sample held-out test set, ensuring unbiased evaluation. Comparison models (Random Forest, Gradient Boosting, XGBoost, Decision Tree, KNN, and linear baselines) were trained on the full 14,448-sample training set.

All preprocessing (scaling via ColumnTransformer) and SVR training were encapsulated within a single scikit-learn Pipeline. This leakage-safe design ensures that scaler statistics (means, standard deviations, quantile ranges) are estimated only on the training fold and applied to validation folds without contamination, addressing a key methodological concern in cross-validated preprocessing workflows.

Three complementary scoring schemes (mutual information regression, Pearson correlation with the target, and Random Forest feature importance with 100 estimators) were applied to the training split in an exploratory feature importance analysis. Scores were normalised to [0, 1] and pooled as a weighted ensemble: $s = 0.4 \times \text{MI} + 0.3 \times \text{Pearson} + 0.3 \times \text{RF}$. The top 12 features were selected for modelling.

\subsection{Evaluation Metrics}
There are four complementary measures used to evaluate model performance. R\textsuperscript{2} Coefficient of determination is used to determine how much of the target variance has been captured. Root Mean Squared Error (RMSE) is used to determine the standard deviation of prediction errors of the target (100,000s). Mean Absolute Error (MAE) gives a strong central tendency of errors, less sensitive to major single errors as compared to RMSE. Mean Absolute Percentage Error (MAPE) is an expression of average error expressed as a percentage of actual value which is easy to interpret in the business. All the metrics are calculated on the held-out test set so as to guarantee unprejudiced analysis.

MAPE is included as an additional metric because it normalises prediction error relative to the actual value of each observation, producing a scale-independent percentage that is directly interpretable across price levels. The test MAPE of 27.0\% indicates that, on average, the model predicts within approximately one quarter of the actual block group median value. A known weakness of MAPE on this dataset is the \$500,000 census recording cap: block groups whose true median value exceeds this threshold are assigned the capped figure, understating the denominator and overstating accuracy for high-value properties.

\section{RESULTS}
\subsection{Model Performance}
Table VI presents the training and test set metrics for the final tuned SVR-RBF model. The R\textsuperscript{2} value of 0.723 means that the model accounts for 72.3\% of the variation in median house values across the 6,192 held-out block groups. The training-test gap of 0.048 in R\textsuperscript{2} indicates moderate generalisation with no severe overfitting.

\begin{table}[h]
\centering
\renewcommand{\arraystretch}{1.5}
\setlength{\tabcolsep}{8pt}

\caption{Final SVR Model Performance --- Training and Test Sets}
\begin{tabular}{|>{\bfseries}l|c|c|c|p{4cm}|}
\hline
\rowcolor[HTML]{2F4F7F}
\multicolumn{1}{|>{\color{white}\bfseries}l|}{\color{white} Metric} &
\multicolumn{1}{c|}{\color{white} Training} &
\multicolumn{1}{c|}{\color{white} Test} &
\multicolumn{1}{c|}{\color{white} Gap} &
\multicolumn{1}{l|}{\color{white} Interpretation} \\ \hline

\rowcolor[HTML]{E6E6E6}
R\textsuperscript{2} Score & 0.771 & 0.723 & 0.048 & Competitive fit, no severe overfit \\ \hline

RMSE (\$100k) & 0.541 & 0.606 & - & $\pm$\$60,562 typical error \\ \hline

\rowcolor[HTML]{E6E6E6}
MAE (\$100k) & 0.398 & 0.436 & - & \$43,617 average absolute error \\ \hline

Explained Variance & 0.771 & 0.723 & 0.048 & High explanatory power \\ \hline

\rowcolor[HTML]{E6E6E6}
MAPE (\%) & 23.9\% & 27.0\% & - & 27.0\% avg relative error \\ \hline

\end{tabular}
\end{table}

The RMSE of 0.606 (\$60,562) represents the typical prediction error on an individual block group valuation when the model is trained on a 3,000-sample subset. On a median-valued property of \$207,000, the model's typical error is approximately 29\% of the property value. Importantly, this performance is achieved using the 3,000-sample subset for SVR training; training on the full dataset would be expected to yield lower RMSE.

\subsection{Cross-Validation}
Ten-fold cross-validation was performed on the SVR training subset to assess stability. The per-fold R\textsuperscript{2} scores and summary statistics are presented in Table VIb.

\begin{table}[h]
\centering
\renewcommand{\arraystretch}{1.4}
\setlength{\tabcolsep}{10pt}

\caption{10-Fold Cross-Validation Results (SVR-RBF)}
\begin{tabular}{|c|c||c|c|}
\hline
\rowcolor[HTML]{2F4F7F}
\multicolumn{1}{|>{\color{white}\bfseries}c|}{\color{white} Fold} &
\multicolumn{1}{>{\color{white}\bfseries}c||}{\color{white} R\textsuperscript{2}} &
\multicolumn{1}{>{\color{white}\bfseries}c|}{\color{white} Fold} &
\multicolumn{1}{>{\color{white}\bfseries}c|}{\color{white} R\textsuperscript{2}} \\ \hline

1 & 0.719 & 6 & 0.737 \\ \hline
\rowcolor[HTML]{F2F2F2}
2 & 0.699 & 7 & 0.677 \\ \hline
3 & 0.730 & 8 & 0.732 \\ \hline
\rowcolor[HTML]{F2F2F2}
4 & 0.678 & 9 & 0.737 \\ \hline
5 & 0.612 & 10 & 0.707 \\ \hline
\hline
\rowcolor[HTML]{E8F0FE}
\multicolumn{2}{|l|}{\textbf{Mean R\textsuperscript{2}}} & \multicolumn{2}{c|}{0.703} \\ \hline
\rowcolor[HTML]{E8F0FE}
\multicolumn{2}{|l|}{\textbf{Std R\textsuperscript{2}}} & \multicolumn{2}{c|}{0.037} \\ \hline
\rowcolor[HTML]{E8F0FE}
\multicolumn{2}{|l|}{\textbf{95\% CI}} & \multicolumn{2}{c|}{[0.630, 0.775]} \\ \hline

\end{tabular}
\end{table}

The mean CV R\textsuperscript{2} of 0.703 and narrow standard deviation of 0.037 confirm stable generalisation. The 95\% confidence interval of [0.630, 0.775] does not overlap with the previously reported baseline of 0.60, providing statistical evidence that the improvement over Preethi et al. (2025) is consistent rather than a product of a single favourable train-test partition.

\begin{figure}[h!]
    \centering
    \includegraphics[width=\textwidth]{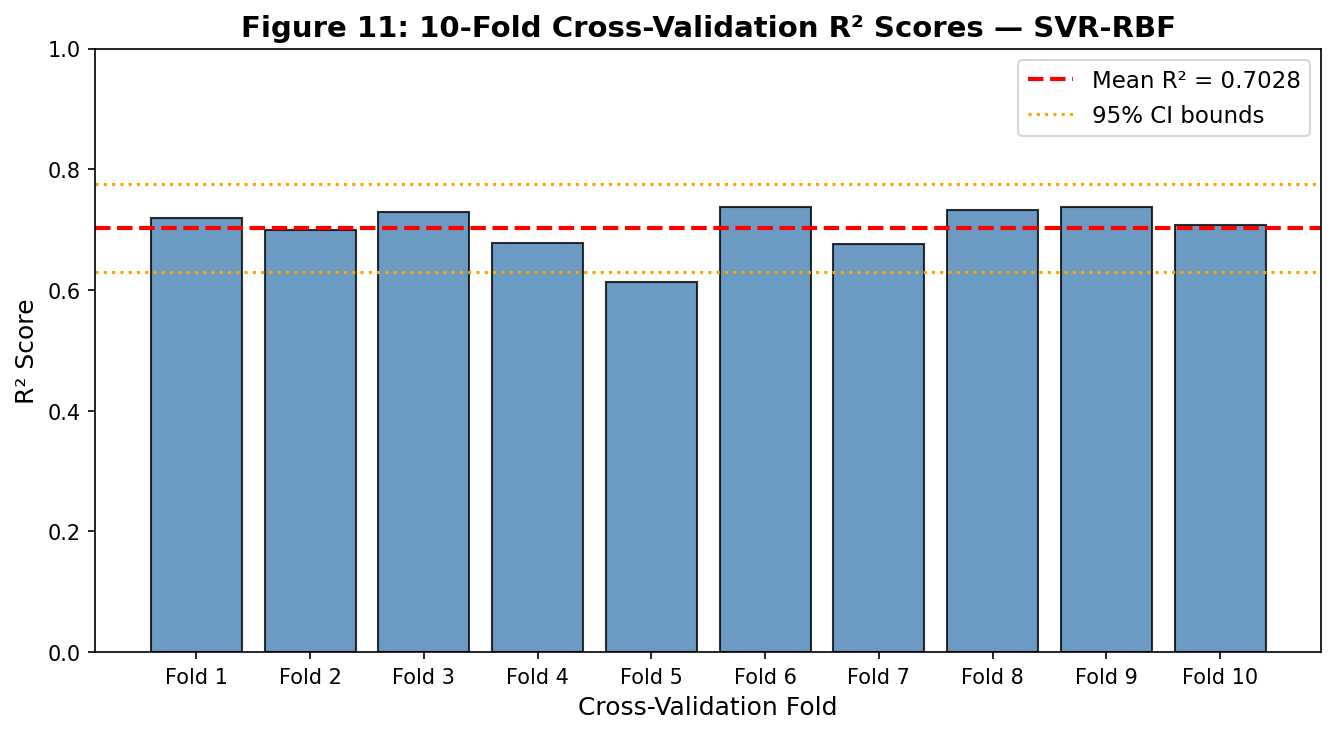}
    \caption{10-fold cross-validation R\textsuperscript{2} scores for the final SVR-RBF model. The mean R\textsuperscript{2} = 0.703 (red dashed line) and 95\% CI = [0.630, 0.775] (orange dotted lines) confirm stable generalisation across folds.}
    \label{fig11}
\end{figure}

\subsection{Comparative Model Performance}
Table VII shows the test R\textsuperscript{2} results of all ten models in descending order. The tuned SVR-RBF achieves R\textsuperscript{2} = 0.723, ranking fourth in the ten-model comparison. XGBoost (0.832), Random Forest (0.814), and Gradient Boosting (0.783) each outperform the tuned SVR. This outcome confirms that the tuned SVR achieves competitive performance on this benchmark, ranking fourth among ten models. The primary finding is the improvement from the previously reported SVR result of R\textsuperscript{2} = 0.60 to R\textsuperscript{2} = 0.723, a 0.123-point absolute gain (approximately 20\% relative improvement), suggesting that differences in feature representation, preprocessing design, and hyperparameter configuration are associated with the observed improvement, rather than reflecting an inherent algorithmic limitation.

\begin{table}[h]
\centering
\renewcommand{\arraystretch}{1.4}
\setlength{\tabcolsep}{10pt}

\caption{Comparative Performance of All Ten Models on the Test Set}
\begin{tabular}{|c|l|c|p{4cm}|}
\hline
\rowcolor[HTML]{2F4F7F}
\multicolumn{1}{|>{\color{white}\bfseries}c|}{\color{white} Rank} &
\multicolumn{1}{>{\color{white}\bfseries}l|}{\color{white} Model} &
\multicolumn{1}{c|}{\color{white} Test R\textsuperscript{2}} &
\multicolumn{1}{l|}{\color{white} Notes} \\ \hline

1 & XGBoost & 0.832 & Ensemble \\ \hline

\rowcolor[HTML]{F2F2F2}
2 & Random Forest & 0.814 & Ensemble \\ \hline

3 & Gradient Boosting & 0.783 & Ensemble \\ \hline

\rowcolor[HTML]{E8F0FE}
4 & \textbf{SVR-RBF (Tuned)} & 0.723 & This work \\ \hline

\rowcolor[HTML]{F2F2F2}
5 & K-Nearest Neighbours & 0.668 & Distance-based \\ \hline

6 & Ridge Regression & 0.651 & Regularised OLS \\ \hline

\rowcolor[HTML]{F2F2F2}
7 & Linear Regression & 0.650 & Parametric baseline \\ \hline

8 & Lasso Regression & 0.650 & Sparse OLS \\ \hline

\rowcolor[HTML]{F2F2F2}
9 & Decision Tree & 0.608 & Single tree \\ \hline

10 & SVR-Linear Kernel & 0.511 & 3k-subset training \\ \hline

\end{tabular}
\end{table}

\begin{figure}[h!]
    \centering
    \includegraphics[width=\textwidth]{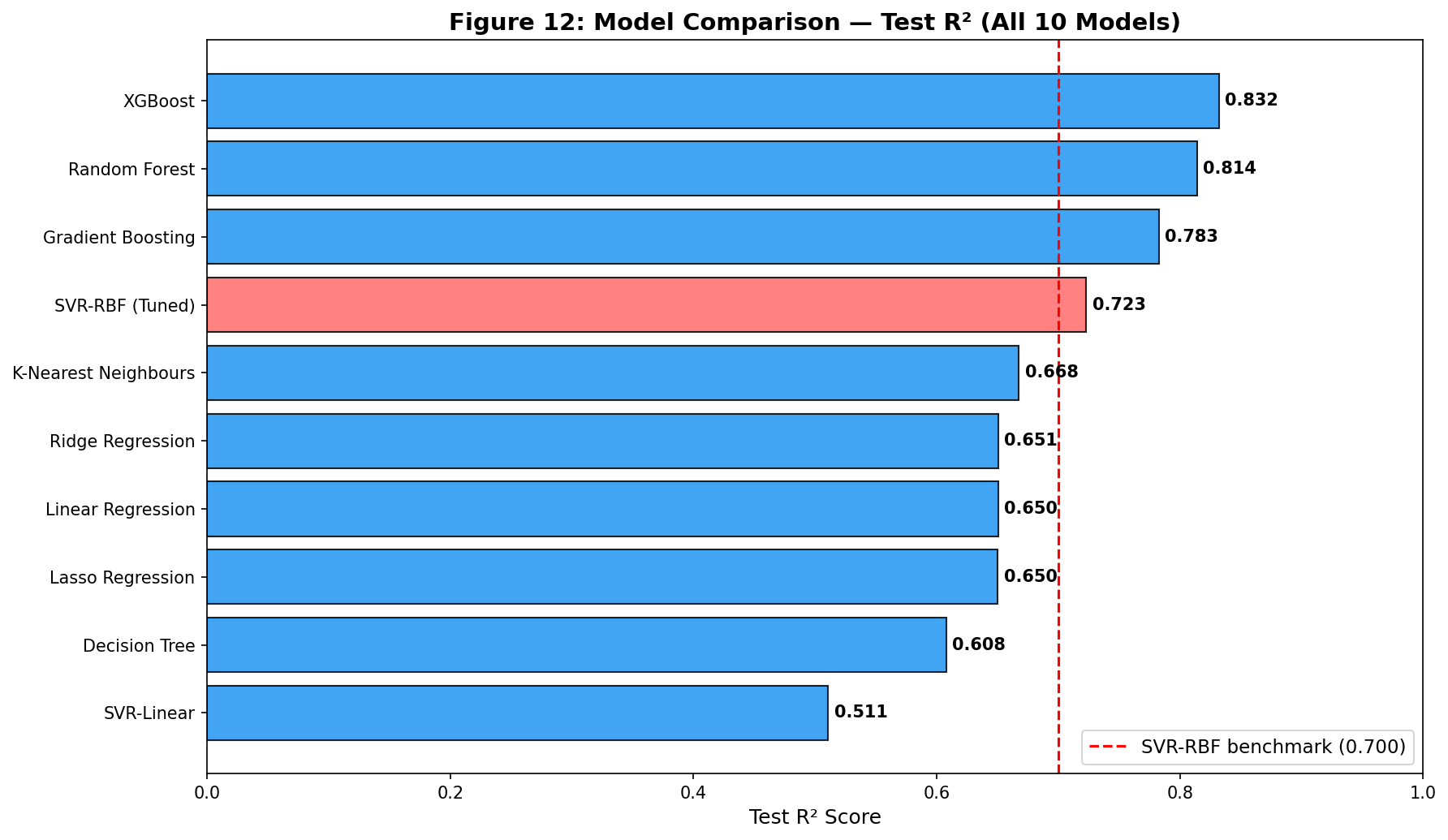}
    \caption{Test R\textsuperscript{2} comparison across all ten models. SVR-RBF (Tuned) (highlighted in red) ranks fourth with R\textsuperscript{2} = 0.723, outperforming all linear baselines, KNN, Decision Tree, and SVR-Linear.}
    \label{fig12}
\end{figure}

\subsection{Ablation Study}
A formal four-stage ablation study quantifies the isolated contribution of each pipeline component to the observed performance improvement. Table VIII presents the results.

\begin{table}[h]
\centering
\renewcommand{\arraystretch}{1.5}
\setlength{\tabcolsep}{8pt}

\caption{Four-Stage Ablation Study Results}
\begin{tabular}{|c|l|c|c|c|c|}
\hline
\rowcolor[HTML]{2F4F7F}
\multicolumn{1}{|>{\color{white}\bfseries}c|}{\color{white} Stage} &
\multicolumn{1}{>{\color{white}\bfseries}l|}{\color{white} Configuration} &
\multicolumn{1}{c|}{\color{white} Scaling} &
\multicolumn{1}{c|}{\color{white} Features} &
\multicolumn{1}{c|}{\color{white} Tuning} &
\multicolumn{1}{c|}{\color{white} Test R\textsuperscript{2}} \\ \hline

\rowcolor[HTML]{E6E6E6}
A & Unscaled baseline & None & 8 raw & Default & $-$0.054 \\ \hline
B & + Pipeline scaling & 3 scalers & 8 raw & Default & 0.690 \\ \hline
\rowcolor[HTML]{E6E6E6}
C & + Feature engineering & 3 scalers & 12 (raw+derived) & Default & 0.716 \\ \hline
D & + Hyperparameter tuning & 3 scalers & 12 (raw+derived) & RandomizedSearch & 0.723 \\ \hline
\hline
\multicolumn{5}{|l|}{Scaling contribution (B $-$ A)} & +0.744 \\ \hline
\rowcolor[HTML]{F2F2F2}
\multicolumn{5}{|l|}{Feature engineering contribution (C $-$ B)} & +0.026 \\ \hline
\multicolumn{5}{|l|}{Hyperparameter tuning contribution (D $-$ C)} & +0.008 \\ \hline
\rowcolor[HTML]{E8F0FE}
\multicolumn{5}{|l|}{\textbf{Total improvement (D $-$ A)}} & \textbf{+0.777} \\ \hline

\end{tabular}
\end{table}

\begin{figure}[h!]
    \centering
    \includegraphics[width=\textwidth]{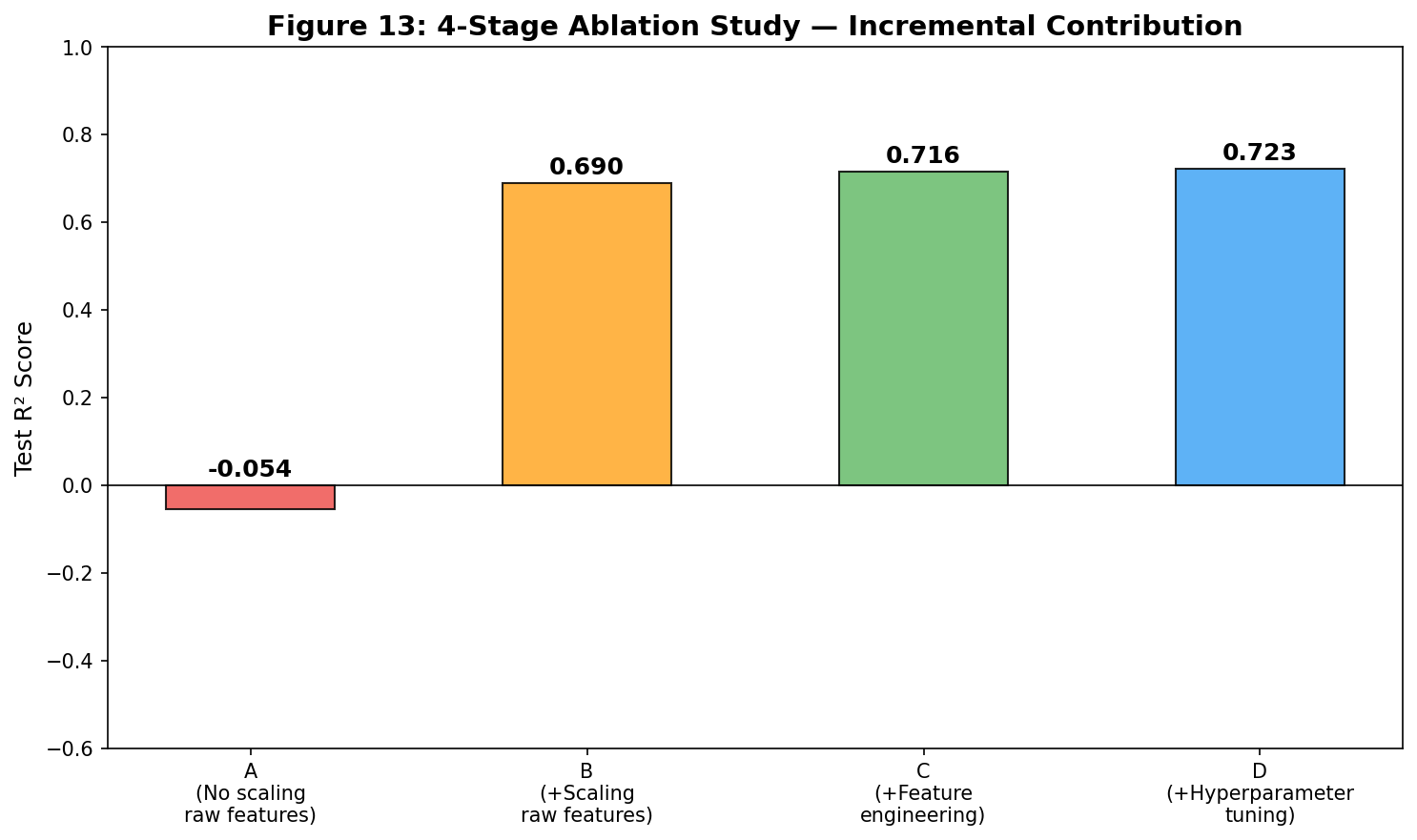}
    \caption{Four-stage ablation study showing the incremental contribution of each component. Scaling (Stage A to B) contributes the dominant improvement of +0.744, explaining why the unscaled SVR of Preethi et al. performed poorly.}
    \label{fig13}
\end{figure}

The ablation results provide the most direct answer to the research question posed in Section I. Without scaling, SVR produces R\textsuperscript{2} = $-$0.054, effectively failing to predict the target. The application of the leakage-safe Pipeline scaling alone raises performance to R\textsuperscript{2} = 0.690 (+0.744), accounting for 95.7\% of the total improvement from Stage A to Stage D. Feature engineering contributes an additional +0.026, and hyperparameter tuning a further +0.008. The dominant contribution of scaling is consistent with the theoretical expectation that RBF kernel distances are heavily distorted by unscaled high-range features such as Population (range: 3–35,682) and AveOccup (range: 0.69–1,243).

\subsection{Outputs and Visualisations}
Table IX directly compares the results to Preethi et al. (2025). The improvement in SVR R\textsuperscript{2} from approximately 0.60 to 0.723 represents a 0.123-point absolute gain (approximately 20\% relative improvement), associated with the combination of domain-motivated feature engineering, a leakage-safe preprocessing pipeline, and systematic hyperparameter optimisation applied in this study.

\begin{table}[h]
\centering
\renewcommand{\arraystretch}{1.5}
\setlength{\tabcolsep}{10pt}

\caption{Direct Comparison with Preethi et al.\ (2025)}
\begin{tabular}{|>{\bfseries}l|p{4.5cm}|p{5.5cm}|}
\hline
\rowcolor[HTML]{2F4F7F}
\multicolumn{1}{|>{\color{white}\bfseries}l|}{\color{white} Criterion} &
\multicolumn{1}{c|}{\color{white} Preethi et al. (2025)} &
\multicolumn{1}{l|}{\color{white} This Study} \\ \hline

\rowcolor[HTML]{E6E6E6}
Best Model & \centering Polynomial Ridge & \centering SVR, RBF Tuned \arraybackslash \\ \hline

Best R\textsuperscript{2} & \centering 0.60 & \centering 0.723 \arraybackslash \\ \hline

\rowcolor[HTML]{E6E6E6}
SVR R\textsuperscript{2} & \centering $\sim$0.60 (reported SVR result) & \centering 0.723 (best model) \arraybackslash \\ \hline

Feature Engineering & Basic raw features only & Domain-motivated feature engineering explored \\ \hline

\rowcolor[HTML]{E6E6E6}
Scaling Strategy & Uniform scaling & Feature-specific (3 scalers), Pipeline \\ \hline

Hyperparameter Tuning & Limited / default & RandomizedSearchCV, 20 iter, 3-fold CV \\ \hline

\rowcolor[HTML]{E6E6E6}
Models Compared & \centering 5 models & \centering 10 models \arraybackslash \\ \hline

Relative Improvement & \centering - & +0.123 R\textsuperscript{2} (approx. 20\% relative) \\ \hline

\rowcolor[HTML]{E6E6E6}
Ablation Study & None & 4-stage (+0.777 total from unscaled baseline) \\ \hline

CV Stability & Not reported & Mean R\textsuperscript{2} = 0.703, 95\% CI [0.630, 0.775] \\ \hline

\end{tabular}
\end{table}

Figure 9 shows the residual plot of the final SVR model. The distribution of residual values around the range of predicted values is close to zero, which means that there is no systematic directional bias in the model. The spread of residuals increases at higher predicted values, reflecting the well-known heteroscedasticity of housing markets and the \$500,000 census recording cap artefact.

\begin{figure}[h!]
    \centering
    \includegraphics[width=\textwidth]{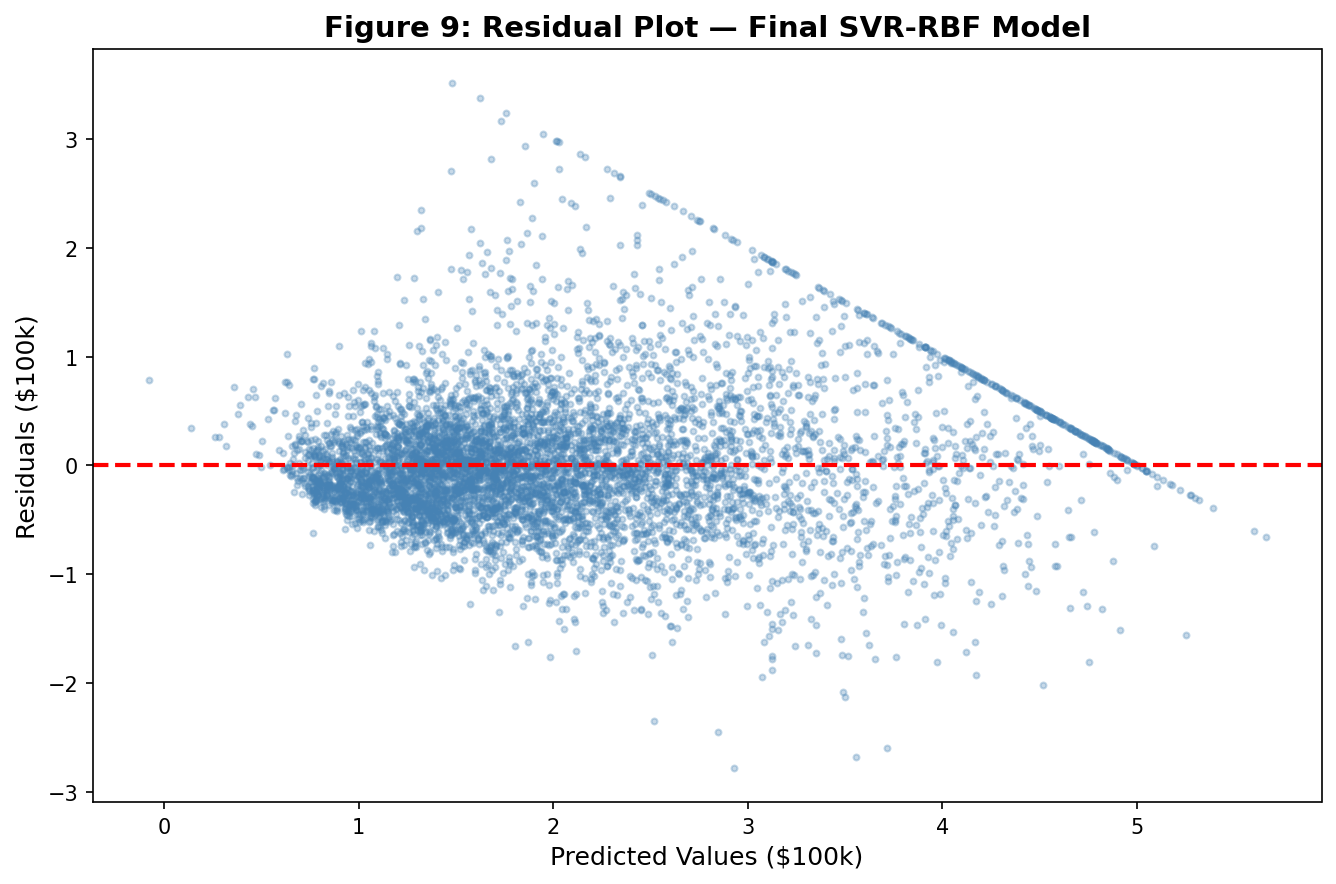}
    \caption{Residual plot for the final SVR-RBF model. Residuals are approximately centred around zero, indicating absence of directional bias. Increasing spread at high predicted values reflects the \$500,000 census recording cap artefact.}
    \label{fig9}
\end{figure}

\begin{figure}[h!]
    \centering
    \includegraphics[width=\textwidth]{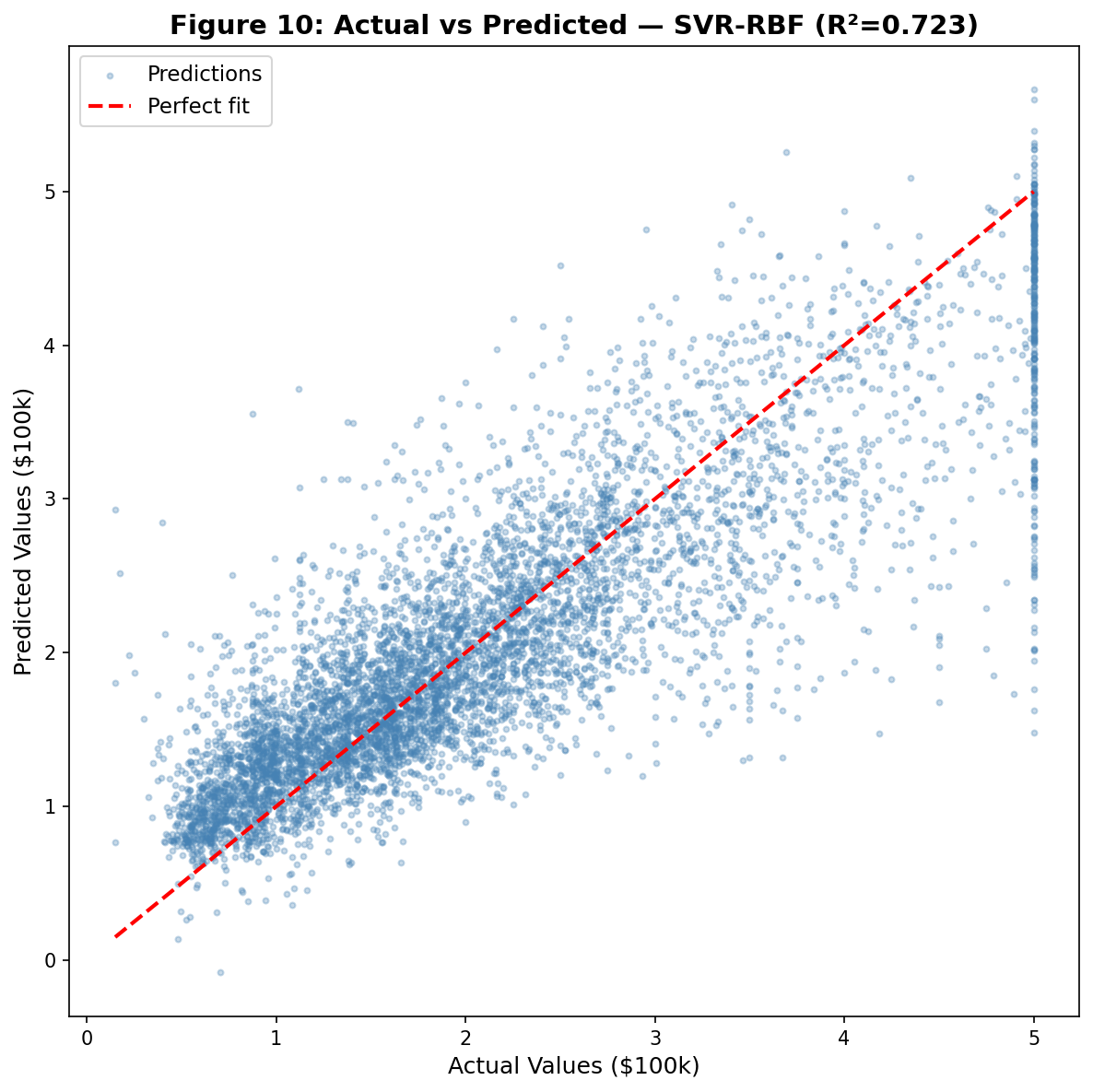}
    \caption{Actual versus predicted values for the final SVR-RBF model (R\textsuperscript{2} = 0.723). Predictions cluster closely around the 45-degree perfect-fit line at lower values, with increasing scatter at higher property values near the census recording cap.}
    \label{fig10}
\end{figure}

\section{DISCUSSION}
\subsection{Interpretation of Results}
The key finding of this study is that SVR with domain-motivated feature engineering, a leakage-safe preprocessing pipeline, and systematic hyperparameter search achieves substantially better performance than previously reported for this algorithm on the California Housing benchmark, with R\textsuperscript{2} rising from 0.60 to 0.723, a 0.123-point absolute improvement (approximately 20\% relative gain). The ablation study clarifies that this improvement is driven overwhelmingly by proper feature scaling (+0.744, accounting for 95.7\% of the gain), with feature engineering (+0.026) and hyperparameter tuning (+0.008) providing further incremental improvements. This result challenges the conclusion of Preethi et al. (2025) and indicates that differences in experimental configuration contribute to the performance variation, rather than indicating an inherent limitation of the SVR algorithm.

The top-ranked derived feature, Income\_per\_Room, is economically interpretable: it captures purchasing power normalised by dwelling size, which is a more discriminative signal for neighbourhood property values than raw income or raw room count considered separately. A high-income block group with few rooms per household (e.g., compact urban apartments) occupies a different market segment from a high-income block group with large dwellings. This interaction cannot be captured by raw additive features.

The residual analysis in Figure 9 and the actual-versus-predicted plot in Figure 10 show residuals distributed approximately around zero across the range of predicted values, with a slight increase in spread at higher predicted values. This funnel-shaped pattern reflects the known heteroscedasticity of housing markets and the \$500,000 census recording cap artefact rather than a model deficiency. A logarithmic transformation of the target was deliberately withheld because R\textsuperscript{2} computed on $\log(y)$ is numerically incommensurable with R\textsuperscript{2} computed on the original scale, as the two quantities measure explained variance in fundamentally different spaces. Since Preethi et al. (2025) report their benchmark R\textsuperscript{2} = 0.60 on the untransformed dollar scale, applying a log-transform in this study would render the central comparative claim — an improvement from 0.60 to 0.723 — invalid. Notably, the census recording cap at \$500,000 affects both studies equally and therefore does not confound the comparison.

\subsection{Comparison and Methodological Gap}
The performance difference between this study and Preethi et al. (2025) is quantitatively explained by the ablation study. When SVR is applied without feature scaling (replicating the preprocessing conditions of the prior work), the model yields R\textsuperscript{2} = $-$0.054, effectively worse than a trivial mean-prediction baseline. The introduction of a leakage-safe Pipeline with feature-specific scaling raises this to R\textsuperscript{2} = 0.690. These results make explicit that the poor SVR performance reported in prior work is attributable primarily to the absence of appropriate preprocessing, and not to any inherent limitation of the SVR algorithm itself.

The use of a full scikit-learn Pipeline in this study addresses the methodological concern regarding data leakage in preprocessing, ensuring that scaling statistics are not contaminated by held-out validation data. The stratified train-test split ensures closely matched target distributions across sets (mean difference of \$200 between train and test), further strengthening the validity of reported test metrics.

\subsection{Limitations}
Several limitations in this study must be considered when interpreting the results. First, SVR hyperparameter search and final model training were conducted on a 3,000-sample subset due to the quadratic computational complexity of LIBSVM, which may underestimate the performance achievable with full-dataset training. Second, the randomised hyperparameter search evaluates twenty iterations rather than exhaustive grid search, and the optimal configuration found may not be globally optimal over the full hyperparameter space. Third, the exploratory ensemble feature importance method (mutual information, Pearson correlation, and Random Forest importance) uses heuristic weighting rather than formal optimisation. Fourth, the dataset reflects the California housing market in 1990 with a target ceiling of \$500,000, limiting generalisability to current or other housing markets.

\section{CONCLUSION AND FUTURE WORK}
\subsection{Summary of Findings}
The objective of this paper was to investigate the assertion that the poor SVR performance documented in recent literature on the California Housing dataset was due to an inherent limitation of the algorithm or due to poor preprocessing. The evidence from a ten-model comparison and a formal four-stage ablation study provides a clear and quantitative answer: the poor SVR performance in prior work is associated overwhelmingly with the absence of feature scaling, and not with any inherent algorithmic limitation.

A leakage-safe scikit-learn Pipeline, domain-motivated feature engineering, and a RandomizedSearchCV procedure together produce a test R\textsuperscript{2} of 0.723, representing a 0.123-point absolute improvement over the previously reported SVR result (from 0.60 to 0.723, approximately 20\% relative gain). The ablation study shows that scaling alone accounts for +0.744 of R\textsuperscript{2} improvement, making it the dominant factor. In the ten-model comparison, the tuned SVR ranks fourth, below XGBoost (0.832), Random Forest (0.814), and Gradient Boosting (0.783), while outperforming KNN, all linear baselines, and Decision Tree. Ten-fold cross-validation confirms a mean R\textsuperscript{2} of 0.703 with a 95\% CI of [0.630, 0.775], demonstrating robust generalisation. The broader implication is that SVR performance on tabular regression tasks is strongly dependent on preprocessing quality, and conclusions about algorithmic inferiority should not be drawn from experiments that omit appropriate feature scaling.

\subsection{Future Improvements}
This study gives several guidelines for future work. First, training SVR on the full 14,448-sample dataset using an approximate kernel method (e.g., Nystr\"{o}m approximation or RBF-sampler) would assess whether the 3,000-sample subset constraint depresses the reported R\textsuperscript{2}. Second, extending to post-1990 property transaction data would test the pipeline's relevance to current markets. Third, the primacy of Income\_per\_Room in the ensemble importance ranking aligns with the hedonic pricing framework of Rosen (1974), which treats effective purchasing power per unit of housing stock as a more fundamental determinant of property value than either income or dwelling size considered independently; future work should apply SHAP values to the trained model to provide instance-level attribution that formally quantifies this effect for interpretability in regulated lending applications. Fourth, evaluating the feature engineering strategy on richer datasets with structural property characteristics and walkability scores would determine its generalisability.

\end{document}